\documentclass[letterpaper, 10 pt, conference]{ieeeconf}  

\IEEEoverridecommandlockouts                              

\usepackage{graphics} 
\usepackage{epsfig} 
\usepackage{mathptmx} 
\usepackage{times} 
\usepackage{amsmath} 
\usepackage{amssymb}  
\usepackage{cite}
\usepackage{multirow}
\usepackage{color}
\usepackage{xcolor}
\usepackage{siunitx}
\usepackage{algorithm}
\usepackage{algorithmic}
\usepackage{booktabs}
\usepackage{stfloats}
\usepackage{bm}
\usepackage{mathrsfs}
\usepackage{makecell}
\usepackage{amsmath}
\makeatletter
\renewcommand{\maketag@@@}[1]{\hbox{\m@th\normalsize\normalfont#1}}%
\makeatother

\title{\LARGE \bf GS-EVT: Cross-Modal Event Camera Tracking \\ based on Gaussian Splatting}

\author{Tao Liu, Runze Yuan, Yi'ang Ju, Xun Xu, Jiaqi Yang, Xiangting Meng, Xavier Lagorce, Laurent Kneip
\thanks{Mobile Perception Lab, ShanghaiTech University}
}

\begin{document}

\maketitle
\thispagestyle{empty}
\pagestyle{empty}

\begin{abstract}

Reliable self-localization is a foundational skill for many intelligent mobile platforms. This paper explores the use of event cameras for motion tracking thereby providing a solution with inherent robustness under difficult dynamics and illumination. In order to circumvent the challenge of event camera-based mapping, the solution is framed in a cross-modal way. It tracks a map representation that comes directly from frame-based cameras. Specifically, the proposed method operates on top of gaussian splatting, a state-of-the-art representation that permits highly efficient and realistic novel view synthesis. The key of our approach consists of a novel pose parametrization that uses a reference pose plus first order dynamics for local differential image rendering. The latter is then compared against images of integrated events in a staggered coarse-to-fine optimization scheme. As demonstrated by our results, the realistic view rendering ability of gaussian splatting leads to stable and accurate tracking across a variety of both publicly available and newly recorded data sequences.

\end{abstract}

\section{INTRODUCTION}
\label{sec:intro}

The ability to efficiently and accurately track the pose of a platform under a variety of conditions is a crucial ability in many emerging robotics, automation, and intelligence augmentation applications. Depending on size and payload of the platform, the most prominent sensor choices in close-to-market solutions are regular cameras and lidars. However, in order to deal with challenges such as high dynamics and low or varying illumination, dynamic vision sensors---also known as event cameras or the silicon retina~\cite{mahowald91,brandli14}---have recently gained in popularity~\cite{gallego2020event}. Rather than measuring absolute intensities, an event camera reacts asynchronously to relative changes in the logarithmic brightness, and fires per-pixel events whenever such changes surpass a given threshold. Owing to this concept, the event camera is able to sense under highly dynamic or adverse illumination conditions.

Despite its many promising advantages, the event camera remains yet to demonstrate its full potential as a tracking sensor. The reason is that in order to localize any kind of exteroceptive sensor, a reference map of fitting modality needs to be given, and the latter is hard to obtain from event cameras. In the most immediate sense, such a reference map would be given as the result of an event-camera Simultaneous Localization And Mapping (SLAM) framework. A few interesting solutions have already been proposed. Kim et al.~\cite{kim2008simultaneous} propose a method that reconstructs a photometric image from purely rotational motion. Ultimate SLAM~\cite{rosinol18} extracts sparse features from event streams. Chamorro et al.~\cite{chamorro2022event} aim at a line feature-based framework. EVO~\cite{rebecq17} considers events to be generated by high appearance gradients in the scene, and---using the separate mapping framework EMVS~\cite{rebecq18}---maintains a volumetric gradient likelihood field. A similar representation is obtained by Wang et al.~\cite{wang22} through a joint tracking and mapping solution. Finally, the filter-based framework by Kim et al.~\cite{kim16} proposes to optimize photometric depth maps. The argument that event-based 3D reference maps are hard to obtain is supported by the fact that the above listed event camera SLAM frameworks have yet to demonstrate a similar level of maturity than their frame-based counterparts. Instead, event cameras often remain a complementary addition to regular cameras~\cite{rosinol18,hidalgo22}.

\begin{figure}[t]
    \centering
    \includegraphics[width=1\linewidth]{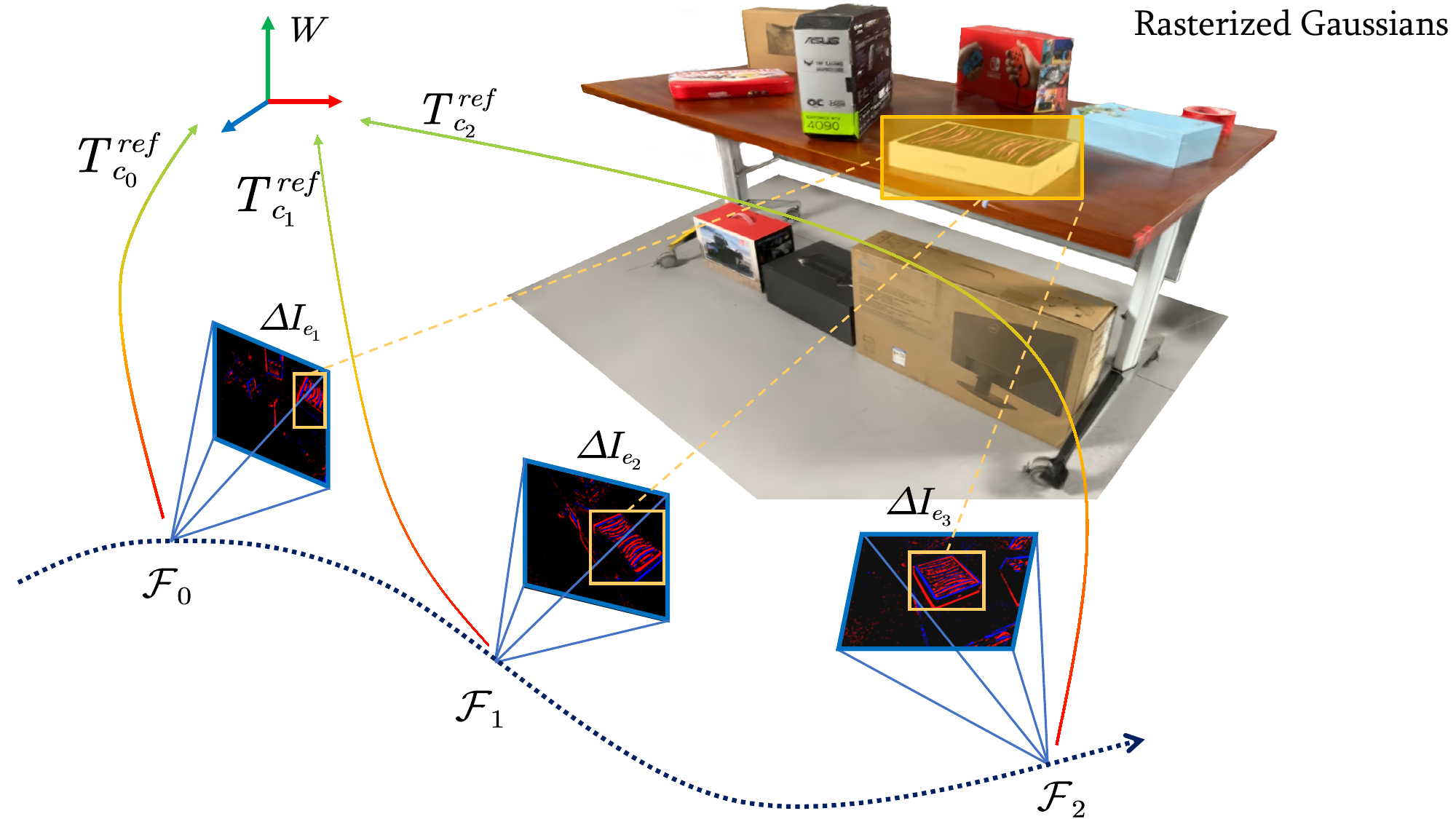}
    \caption{Overview of tracking procedure. The event camera moves 6-DOF freely within the Gaussian map, and high-fidelity intensity images are obtained by rasterizing these Gaussians. We simulate event generation by differentiating between two intensity images. The camera pose is optimized by calculating the photometric error between the intensity changes rendered from the Gaussian map and the actual intensity changes captured by the event camera.}
    \label{fig:enter-label}
\end{figure}

However, the fact that frameworks such as the filtering-based solution by Kim et al.~\cite{kim16} reconstruct a photometric depth map suggests an interesting solution to event camera tracking if an alternative and perhaps more suitable sensor is permitted at mapping time. For example, Gallego et al.~\cite{gallego2017event} and Bryner et al.~\cite{bryner2019event} have already proposed event-camera tracking from photometric depth maps obtained by regular cameras. Yet another example is the works of Zuo et al.~\cite{zuo2022devo,10401990}, who have demonstrated the use of geometric semi-dense maps for event camera tracking, which can be obtained from depth sensors or image-based semi-dense reconstructions. While interesting, both approaches make approximations. Implementations relying on photometric depth maps~\cite{gallego2017event,bryner2019event} promise the ability to simulate events from differentiable novel view synthesis, however make the approximate lambertian surface assumption. Geometric, semi-dense alternatives~\cite{zuo2022devo,10401990} in turn rely on the assumption that events are primarily generated by spatial curves reflecting the 3D location of high appearance gradients.

The motivation of our paper is simple: Use Gaussian Splatting~\cite{kerbl3Dgaussians}, a recent state of the art 3D map representation enabling efficient, differentiable and realistic novel view synthesis for event camera tracking. Similar to NeRF~\cite{mildenhall2021nerf}, Gaussian Splatting models view-dependent radiance of 3D points, and thereby enables more realistic simulation of events. In summary, our contributions are as follows:
\begin{itemize}
  \item To our best knowledge, we proposed the first Gaussian Splatting map-based event-camera tracking framework. The map is hereby used for differential image synthesis, which can be compared against event accumulations to update the pose and velocity of the camera.
  \item We introduce a novel parameterization that employs both the pose of the camera and first-order dynamics. The former is used to optimize the pose of the camera at the center of temporal event accumulation intervals, while the latter is used to enable local differential image rendering based on a constant velocity assumption.
  \item Pose and first-order dynamics are optimized in a staggered scheme that starts with a polarity-free, coarse-to-fine pose-only optimization, and terminates with polarity-aware simultaneous pose and velocity optimization at the finest level.
\end{itemize}

The framework is tested on publicly available benchmark sequences, and---owing to its inherent realistic novel view synthesis ability---outperforms existing photometric methods. Furthermore, our proposed method proves to be more generally applicable, as also the most recent semi-dense geometric alignment frameworks are outperformed in more noisy texture scenarios.

\section{RELATED WORK}
\label{sec:related works}

\subsection{Event-based tracking}
Given a known 3D scene representation and the initial state of the event camera, event-based tracking aims to estimate the camera's trajectory from a continuous stream of events. To achieve this goal, there are generally two approaches: geometric and photometric methods.

When an event camera moves freely within a scene, events are primarily triggered by edges. Geometric methods~\cite{10401990, yuan2024eviteventbasedvisualinertialtracking} optimize the pose of the event camera by aligning 2D events with 3D edge features (2D-3D alignment). For instance, EVT~\cite{10401990} utilizes a 3D curve segment extraction method~\cite{he2018incremental} to build a global semi-dense map, which heavily relies on geometric features in the scene. However, when the scene texture is noisy, the effective extraction of 3D edge features becomes challenging, increasing the likelihood of a tracking loss. Furthermore, semi-dense methods do not effectively handle occlusions and easily cause wrong associations between events and edge features.

The photometric method~\cite{bryner2019event, gallego2017event} captures intensity changes via an appearance-informed 3D scene representation and camera motion. By maximizing the consistency between intensity changes and events, the camera pose is optimized. This approach does not encounter the occlusion problem~\cite{newcombe2011kinectfusion, whelan2015elasticfusion, newcombe2011dtam}. However, photometric maps generally do not provide view-dependent appearance, whereas illumination in the real world does change with different viewpoints.

\subsection{Gaussian Splatting map}
3D Gaussian Splatting (3DGS)~\cite{kerbl3Dgaussians} is an innovative 3D map representation that offers more realistic details compared to semi-dense or dense maps. By using spherical harmonics (SH), the 3DGS primitives can be rendered into different colors from various viewpoints. Thus, the map can encode more complex ambient lighting and aligns more closely with the working principle of event cameras.

In addition to rendering photo-realistic color images, 3DGS also possesses good differentiability, facilitating easy optimization of the sensor state. For instance, by incorporating 3DGS with SLAM~\cite{matsuki2024gaussian, keetha2024splatam, yan2024gs, yugay2023gaussian, peng2024rtg, huang2024photo, ha2024rgbd}, the camera poses and 3DGS are incrementally optimized with each incoming frame, which is considered as an online process. In contrast, some representative offline methods~\cite{fan2024instantsplat, fu2024colmapfree3dgaussiansplatting} jointly optimize the initialized poses and 3DGS map in a single step, aiming for a comprehensive refinement of all frames at once.

The above works provide a practical solution for computing the derivative of poses w.r.t. 3DGS. A key source of inspiration for our work is MonoGS~\cite{matsuki2024gaussian}, which provides the first analytical Jacobian of the camera pose. We extend this methodology by incorporating velocity optimization as well, aiming to further enhance the robustness of our system.

\section{METHODOLOGY}
\label{sec:methodology}

\subsection{Intensity change image from event camera}
\label{sec:event_camera}

\noindent\textbf{Event camera:}
Unlike traditional frame-based cameras that capture images at a fixed frequency, event cameras operate in a quasi-continuous temporal stream-based fashion. Each pixel in an event camera has an independent circuit that responds asynchronously to changes in relative contrast. If $I(\mathrm{u}_k, \tau_k)$ represents the intensity detected at the pixel $\mathrm{u}_k = (x_k, y_k)$ corresponding to timestamp $\tau_k$, an event is generated when the intensity change exceeds a specified threshold $C$:
\begin{equation}
\label{eq:delta_Ie}
    \Delta \ln I_e(\mathrm{u}_k, \tau_k) \doteq \ln I_e(\mathrm{u}_k, \tau_k) - \ln I_e(\mathrm{u}_k, \tau_k - \Delta \tau_k) = p_k C
\end{equation}
where $C$ is called the contrast sensitivity, $p_k = \{-1, +1\}$ indicates the polarity of events, and $\Delta \tau_k$ denotes the time interval since the last event was recorded at the same pixel.

\noindent\textbf{Adaptive event keyframe generation:}
The temporal rate at which events are triggered exhibits a strong correlation with both the current texture of the scene and the velocity of the camera. Therefore, previous works~\cite{9386209, zuo2022devo} that form keyframes at a fixed rate are susceptible to high dynamics and often result in suboptimal optimization outcomes. Specifically, fixed-rate keyframe formation results in unnecessary frames when there is minimal motion, or causes inadequate frames violating the principle that they represent local information during rapid movements. 
Towards adaptive keyframe determination, we process events in groups $\epsilon = \{e_k\}_{k=1}^N$ and generate new event keyframes whenever a minimum number of events has been achieved. These events are accumulated pixel-wise in the image plane, producing an image $\Delta I_e(\mathrm{u}, \tau)$ that indicates the intensity change during a time interval:
\begin{equation}
    \Delta I_e(\mathrm{u}, \tau) = \sum\limits_{k \in N} \Delta \ln I_e(\mathrm{u}_k, \tau_k),
\end{equation}
where the timestamp $\tau$ of the keyframe is defined as the midpoint between the first and last timestamp of the accumulated events in the image plane, $\mathrm{u}$ is the collection of all pixels, and $\Delta I_e$ represents the (logarithmic) intensity change image from the event camera.

\begin{figure*}[t]
    \centering
    \includegraphics[width=1\linewidth]{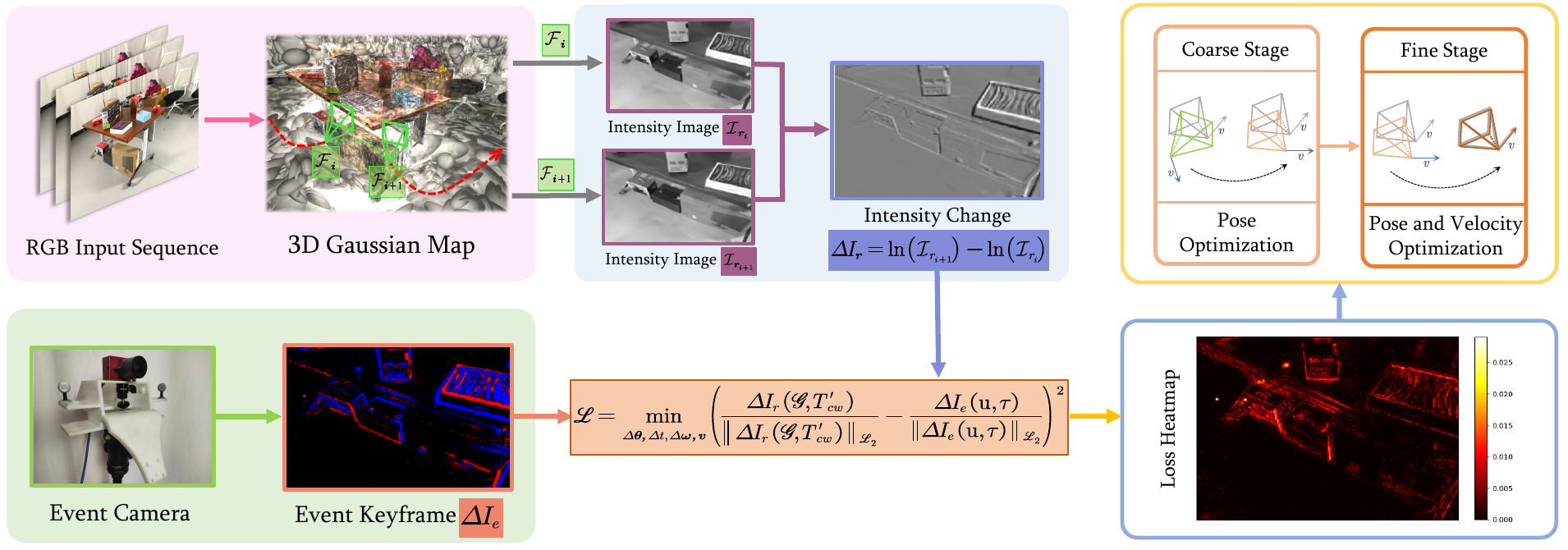}
    \caption{Block diagram of GS-EVT pipeline. We build the 3DGS map from a sequence of RGB images. Within this map, the event camera accumulates a keyframe over a short trajectory (green line). The poses of the first and last events (green frustum) within the keyframe are determined in Eq~\ref{eq:T_first_last}. Using these poses, we render two intensity images from the map. The difference between these images generates a rendered intensity change image. The event keyframe is directly accumulated from the event stream. By utilizing a photometric loss function, we initially perform a polarity-free pose-only optimization in the coarse stage, followed by a refinement of the pose incorporating velocity optimization in the fine stage.}
    \label{fig:pipeline}
\end{figure*}

\subsection{Intensity change image from Gaussian Splatting}
\label{sec:gaussian_splatting}

\noindent\textbf{Gaussian Splatting:}
In contrast to traditional 3D scene representations such as meshes or point clouds, 3DGS utilizes spherical harmonics (SH) to capture the view-dependent appearance of a scene. This allows for a more realistic depiction of lighting changes in the real world and is particularly well-suited for interacting with event cameras, whose working principle relies on sensing intensity change to generate events. 

A 3D Gaussian can be described using its mean $\mu \in \mathbb{R}^3$, covariance matrix $\Sigma \in \mathbb{R}^{3 \times 3}$, color $c \in \mathbb{R}^3$ and opacity $o \in \mathbb{R}$. To render an image, we first project the 3D Gaussians into the 2D image plane:
\begin{equation}
\label{eq:mu_and_sigma}
    \mu^{\prime} = \Pi(T_{cw}\mu), \quad \Sigma^{\prime}=JR_{cw}\Sigma{R_{cw}}^{\top}J^\top,
\end{equation}
where $\mu^{\prime}$ is the mean of the 2D Gaussian, $T_{cw} \in SE(3)$ represents the transformation from world coordinates to camera coordinates, and $\Pi$ defines the projection operation. Since the perspective projection of a 3D Gaussian does not yield a 2D Gaussian, we use the affine transformation $J\in \mathbb{R}^{3 \times 3}$ from~\cite{zwicker2001surface} to approximately get $\Sigma^{\prime}$.

The visible 2D Gaussians are then sorted according to their depth, and front-to-back $\alpha$-blending is applied to render the color of every pixel:
\begin{equation}
C_{i}=\sum_{n\leq{N}}c_{n} \cdot \alpha_{n} \cdot \prod\limits_{m=1}^{n-1}(1-\alpha_{m}),
\end{equation}
where $c_n$ denotes the color of a Gaussian defined by spherical harmonics, and $\alpha_n$ refers to its density, which is the product of the 2D covariance $\Sigma^\prime$ and the opacity $o$.

\noindent\textbf{Generating the intensity change image:}
Instead of relying on depth information for optical flow calculation to predict intensity changes~\cite{bryner2019event}, our method requires only a monocular camera to construct a 3DGS map. The intensity change image is then generated by calculating the difference between rendered images from two different poses. The question is how these poses are defined as a function of position, orientation, and first-order dynamics.

\begin{figure}[t]
    \centering
    \includegraphics[width=0.43\textwidth]{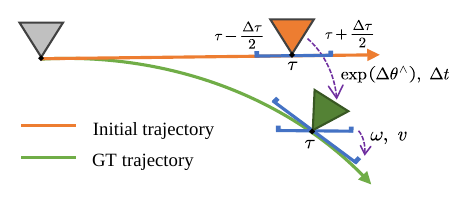}
    \caption{According to the constant velocity model, we estimate the initial camera pose (orange frustum) based on the previous pose (gray frustum). The blue line represents the approximated keyframe accumulation trajectory, which spans a duration of $\Delta \tau$ time. $\tau - \frac{\Delta \tau}{2}$ and $\tau + \frac{\Delta \tau}{2}$ denote the first and last event timestamps within that interval. By applying pose optimization ($\mathrm{exp}((\Delta\theta)^{\wedge}), \Delta t$), the initial pose will be adjusted to closely align with the ground truth trajectory. Subsequently, velocity optimization ($\omega, v$) is used to ensure that the keyframe accumulation trajectory is tangent to the GT trajectory (velocity direction) and its length is corrected so that the thickness of rendered event's edge matches the real event's edge (velocity magnitude).}
    \label{fig:trajectory}
\end{figure}

Our key idea consists of using a pose variable to define the keyframe located at the center of the time interval $\Delta\tau$, and velocities to create differential frames at the boundaries of the interval by a relative forward and backward displacement via a constant velocity motion model (cf. Fig.~\ref{fig:trajectory}). According to Section~\ref{sec:event_camera}, the keyframe timestamp is derived as the midpoint of the largest and the smallest event timestamp in the interval. The two poses extrapolated by constant velocity motion model then approximately correspond to the locations where the first and last event in the interval $\Delta\tau$ have been captured. They are given by
\begin{align}
\label{eq:T_first_last}
    T_{first} &= \operatorname{v2t}(v \frac{-\Delta \tau}{2}, \omega \frac{-\Delta \tau}{2}) \cdot T_{cw} \notag \\
    T_{last} &= \operatorname{v2t}(v \frac{\Delta \tau}{2}, \omega \frac{\Delta \tau}{2}) \cdot T_{cw},
\end{align}

where $v$ represents linear velocity, $\omega$ denotes angular velocity, $T_{cw}$ is the pose of keyframe, and $\operatorname{v2t}$ creates a transformation matrix from a 6-dimensional minimal representation of a transformation change (translation and Rodrigues vector). With these two poses obtained, we can employ 3DGS to render two intensity images and simulate the intensity change image using:
\begin{equation}
\label{eq:delta_Ir}
    \Delta I_r(\mathcal{G}, T_{cw}) = \ln I_r(\mathcal{G}, T_{last}) - \ln I_r(\mathcal{G}, T_{first}).
\end{equation}
$\Delta I_r$ is the intensity change image rendered by Gaussians $\mathcal{G}$ from $T_{cw}$ under the velocity $v$ and $\omega$. To correctly model the behavior of the event camera, we take the logarithms of the intensity images similar to Eq. \eqref{eq:delta_Ie}.

\subsection{Optimization}
\label{sec:optimization}
Since the rasterisation of 3DGS is implemented with CUDA, the derivatives of all parameters need to be given for gradient back-propagation. To ensure the consistency and efficiency of the system, we compute the pose and velocity Jacobians explicitly as well. To the best of our knowledge, this work represents the first application of camera velocity optimization in the context of 3DGS. 

We first extend $T_{cw}$ as follows:
\begin{footnotesize}
\begin{equation}
T_{cw}^\prime=
\begin{bmatrix}
\mathrm{exp}((\omega\frac{\Delta\tau}{2})^{\wedge}) &  v\frac{\Delta\tau}{2}) \\
0                                     &   1
\end{bmatrix}
\begin{bmatrix}
\mathrm{exp}((\Delta\theta)^{\wedge}) &  \Delta{t} \\
0                      &   1
\end{bmatrix}
\begin{bmatrix}
R_{cw} &  t_{cw} \\
0      &   1
\end{bmatrix},
\label{eq:Tcw}
\end{equation}
\end{footnotesize}
where $\Delta\theta$ indicates the change in rotation, $\Delta t$ represents the change in translation, and $T_{cw}^\prime$ is the extension of $T_{cw}$ that incorporates pose and velocity optimization variables. The notation $a^{\wedge}$ denotes the skew-symmetric matrix. For clarity, we simplify $T_{cw}^\prime = T_1 \cdot T_2 \cdot T_3$.

Inspired by~\cite{matsuki2024gaussian}, we use Lie algebra to derive the Jacobians. According to Eq. \eqref{eq:mu_and_sigma}, the gradient of $T_{cw}^\prime$ is related to $\mu^\prime$ and $\Sigma^\prime$. Following the chain rules, we can get:
\begin{align}
    & \frac{ \partial \mu^\prime }{ \partial T_{cw}^\prime } = \frac{\partial \mu^\prime}{\partial (T_{cw}^\prime \mu)} \frac{\partial (T_{cw}^\prime \mu)}{\partial T_{cw}^\prime} \notag \\
    & \frac{ \partial \Sigma^\prime }{ \partial T_{cw}^\prime } = \frac{\partial \Sigma^\prime}{\partial J}\frac{\partial J}{\partial (T_{cw}^\prime \mu)} \frac{\partial (T_{cw}^\prime \mu)}{\partial T_{cw}^\prime} + \frac{\partial \Sigma^\prime}{\partial R_{cw}} \frac{\partial R_{cw}}{\partial T_{cw}^\prime}.
\end{align}

\noindent\textbf{Pose optimization:}
The pose optimization is designed as a coarse-to-fine scheme. At the coarse stage, we do not distinguish between positive and negative events; the polarity of any triggered event is treated as $p_k=+1$ and $\Delta I_e \geq 0$. For the rendered intensity change (cf. Eq. \eqref{eq:delta_Ir}), we accordingly take the absolute value of $\Delta I_r$. By doing so, the optimization process can be viewed as the registration of two edge images, preventing the algorithm from getting stuck in locally optimal solutions (We encourage the reader to watch the supplementary video, which visualizes this behavior). At the fine stage, we fall back to the original formulation with $p_k=\{-1, +1\}$ and no absolute value operation, which further optimizes the camera pose.

To optimize the pose, we calculate the derivative of $T_{cw}^\prime\mu$ w.r.t. $T_2$. Since $T_2 \in SE(3)$, we denote the Lie algebra of $T_2$ as $\xi$ and apply a perturbation $\Delta{T_2}=exp(\delta\xi^{\wedge})$ to $T_{cw}^\prime\mu$ and $R_{cw}$ (where $ (A)_{:, i}$ refers to the ith column of the matrix $A$):
\begin{equation}
\frac{\partial(T_{cw}^\prime\mu)}{\partial(\Delta{t}, \Delta\theta)} = 
R_{1}
\begin{bmatrix}
    \mathbf{I}  &  -[R_{2}(R_{3}\mu+t_{3})+t_{2}]^{\wedge}\\
    \mathbf{0}^\mathrm{T}     &  \mathbf{0}^\mathrm{T} 
\end{bmatrix} \notag
\end{equation}
\begin{equation}
    \frac{\partial{R}_{cw}}{\partial(\Delta{t}, \Delta\theta)}= 
    R_{1}
    \begin{bmatrix}
    \mathbf{0} & -[(R_2R_3)_{:, 1}]^{\wedge}&   \\
    \mathbf{0} & -[(R_2R_3)_{:, 2}]^{\wedge}&   \\
    \mathbf{0} & -[(R_2R_3)_{:, 3}]^{\wedge}&   
\end{bmatrix}.
\end{equation}

The full derivation of the pose and velocity Jacobians can be found in the supplementary materials.

\noindent\textbf{Velocity optimization:}
The rendered intensity change image is influenced by the poses of the first and the last accumulated event in a keyframe, predicted using a constant velocity model. Consequently, optimizing velocity is crucial for enhancing the accuracy of the intensity change image and 
makes the keyframe accumulation trajectory more tangent to the ground truth trajectory
(cf. Fig.~\ref{fig:trajectory}).  Intuitively, optimizing the keyframe pose changes the location of the strongest response in the intensity change image, while adjusting the magnitude and orientation of the velocity changes the strength of those responses as well as their orientation along the high-gradient regions.

Similar to pose optimization, we use the disturbance model of Lie algebra to calculate the derivative of $T_{cw}^\prime\mu$ w.r.t. $T_1$:
\begin{equation}
\frac{\partial T_{cw}^\prime\mu}{\partial(v, \omega)}=
\frac{\Delta\tau}{2}
\begin{bmatrix}
    \mathbf{I}  &  -[R_{1}(R_{2}(R_{3}\mu+t_{3})+t_{2})+t_{1}]^{\wedge} \\
    \mathbf{0}^\mathrm{T}     &  \mathbf{0}^\mathrm{T}
\end{bmatrix} \notag
\end{equation}
\begin{equation}
\frac{\partial R_{cw}}{\partial (v, \omega)}=
\frac{\Delta\tau}{2}
\begin{bmatrix}
    \mathbf{0} & -[(R_1R_2R_3)_{:, 1}]^{\wedge}   \\
    \mathbf{0} & -[(R_1R_2R_3)_{:, 2}]^{\wedge}   \\
    \mathbf{0} & -[(R_1R_2R_3)_{:, 3}]^{\wedge}
\end{bmatrix}.
\end{equation}

\noindent\textbf{Loss function:}
From Section~\ref{sec:event_camera} and ~\ref{sec:gaussian_splatting}, we can get $\Delta I_e$ by accumulating events up to a specified count $N$, and obtain $\Delta I_r$ by taking the difference between two rendered images at poses $T_{first}$ and $T_{last}$. Therefore, the objective function can be constructed by minimizing the photometric difference between $\Delta I_r$ and $\Delta I_e$:
\begin{equation}
\mathcal{L} = \mathop{\min}\limits_{\Delta{t},\Delta\theta,v,\omega} \left( \Delta{I_{r}(\mathcal{G}, T_{cw}^\prime)}-\Delta{I_{e}(\mathrm{u}, \tau)} \right)^2.
\end{equation}

However, Eq. \eqref{eq:delta_Ie} reveals that the intensity change perceived by event cameras is influenced by the contrast sensitivity $C$, typically an unknown variable. Therefore, the distribution of $\Delta I_e$ and $\Delta I_r$ might be different. To address this issue, we propose to minimize the difference between normalized (unit-norm) images inspired by~\cite{evangelidis2008parametric}:
\begin{equation}
\mathcal{L} = \mathop{\min}\limits_{\Delta{t},\Delta\theta,v,\omega} \left( \frac{\Delta{I_{r}(\mathcal{G}, T_{cw}^\prime)}}{\Vert\Delta{I_{r}(\mathcal{G}, T_{cw}^\prime)}\Vert_{\mathcal{L}_2}} - \frac{\Delta{I_{e}(\mathrm{u}, \tau)}}{\Vert\Delta{I_{e}(\mathrm{u}, \tau)\Vert_{\mathcal{L}_2}}} \right )^2.
\end{equation}

\noindent\textbf{Convergence:}
In the first half stage of the optimization, our focus is on coarse pose optimization. This step primarily aims to quickly establish a registration between two edge images. Subsequently, in the latter half of the optimization process, velocity optimization is integrated with fine pose optimization (cf. Fig.~\ref{fig:pipeline}). This dual approach is designed to address rapid directional changes and to finely adjust the camera pose on a more detailed scale. We set the slope of the loss as the convergence criterion for both stages.

\section{EXPERIMENTS}
\label{sec:experiments}

\begin{figure}
    \centering
    \includegraphics[width=1\linewidth]{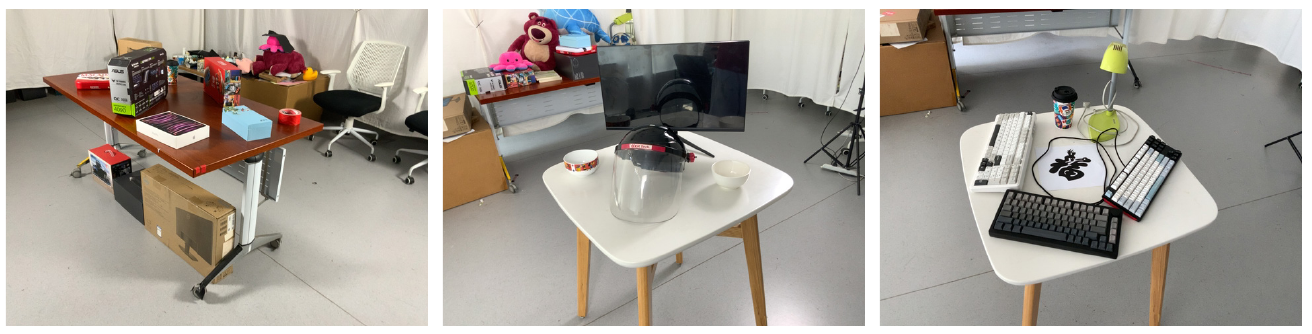}
    \caption{Visualization of self-collected dataset. \emph{desk} sequence with severely occluded scene type (left); \emph{keyboard} sequence complexly textured scene type (middle); \emph{helmet} sequence with highly reflective scene type (right).}
    \label{fig:self-collected-dataset}
\end{figure}

To evaluate the performance of our method, we first conduct experiments on the open-source VECtor dataset~\cite{9809788}. Similar to other SLAM datasets, this dataset does not separately record the trajectory for map-building and tracking. This may be a problem, as for the tracking task, a carefully constructed map is a prerequisite. However, for some of the sequences, the dynamics may be too challenging to build a map and result in the loss of map details or even complete map-building failures. To properly evaluate highly dynamic trajectories on carefully constructed maps, we propose additional self-collected datasets including rich, carefully collected sequences for mapping and separate, more challenging trajectories within the same environment for tracking. We evaluate our approach against state-of-the-art event-based tracking approaches on both datasets.   

\subsection{Implementation details}

We implement several techniques aiming at improving the performance and robustness of our algorithm. First, we apply Gaussian blur to smooth the intensity change images, which helps reduce high-frequency noise and benefits the accuracy of gradient calculations. Following this, we utilize image pyramids to subsample the original image into different resolutions. By starting with a lower resolution, our algorithm can capture the global structure of the images before refining the alignment at higher resolutions, enabling more precise results. To further mitigate the impact of noise from rendered images  and enhance computational efficiency, a mask scheme has been designed. This approach involves initially expanding the region containing events with a certain magnitude. Subsequently, the calculation of the gradient is confined solely within this expanded area to avoid possible noise from artifacts in the 3DGS map.

\subsection{Evaluation on VECtor dataset}

\begin{table}[t]
\caption{Absolute trajectory error (ATE) on VECtor dataset normal sequences. Position: {[}cm{]}, Orientation: {[}$^{\circ}${]} }
\label{tab:vector_dataset}
\resizebox{\columnwidth}{!}{%
\begin{tabular}{@{}ccccccccc@{}}
\toprule
\multirow{2}{*}{Sequence} &
  \multirow{2}{*}{Method} &
  \multirow{2}{*}{Frequency} &
  \multicolumn{2}{c}{30\% sequence} &
  \multicolumn{2}{c}{50\% sequence} &
  \multicolumn{2}{c}{100\% sequence} \\ \cmidrule(l){4-9} 
                              &                      &       & Pos.          & Orient.       & Pos.          & Orient.       & Pos.          & Orient.       \\ \midrule
\multirow{5}{*}{sofa\_normal} & \multirow{3}{*}{EVT} & 300hz & 3.26          & \underline{1.34}          & \underline{3.66}          & \underline{1.75}          & \underline{3.37}          & \underline{1.65}          \\
                              &                      & 100hz & \underline{3.16} & \textbf{1.33} & 6.98          & 2.70          & 5.43          & 2.28          \\
                              &                      & 30hz  & 3.54          & 1.58          & -             & -             & -             & -             \\
                              & EVPT                 & (dyn)   & 15.64         & 5.08          & 15.27         & 5.50          & -             & -             \\
                              & GSEVT                 & (dyn)   & \textbf{3.07}         & 1.54          & \textbf{3.47}         & \textbf{1.72}          & \textbf{3.23}             & \textbf{1.61}             \\
                              \midrule
\multirow{5}{*}{robot\_normal} &
  \multirow{3}{*}{EVT} &
  300hz &
  \textbf{0.91} &
  \underline{0.89} &
  \textbf{1.04} &
  \underline{1.08} &
  \textbf{1.09} &
  \underline{1.31} \\
                              &                      & 100hz & \underline{0.97}          & 0.91          & \underline{1.09}          & 1.09          & \underline{1.16}          & \underline{1.31}          \\
                              &                      & 30hz  & 1.02          & 0.90          & 1.11          & 1.11          & 1.20          & 1.35          \\
                              & EVPT                 & (dyn)  & 15.36         & 9.927         & 13.40         & 8.39          & 14.17         & 9.03          \\
                              & GSEVT                 & (dyn)   & 1.41         & \textbf{0.75}         & 1.60         & \textbf{0.87}          & 2.51             & \textbf{1.27}             \\
                              \midrule
\multirow{5}{*}{desk\_normal} &
  \multirow{3}{*}{EVT} &
  300hz &
  \textbf{1.21} &
  \textbf{0.74} &
  \textbf{1.87} &
  \textbf{0.81} &
  \underline{2.25} &
  \textbf{0.88} \\
                              &                      & 100hz & \underline{1.34}          & \underline{0.77}          & \textbf{1.87}          & \underline{0.84}          & \textbf{2.14} & 0.92          \\
                              &                      & 30hz  & \textbf{1.21} & \underline{0.77}          & \underline{1.92}          & 0.87          & 2.30          & 0.96          \\
                              & EVPT                 &  (dyn) & 7.71          & 1.83          & 11.02         & 2.61          & 16.08         & 3.33          \\
                              & GSEVT                 & (dyn)   & 2.38         & \textbf{0.74}        & 3.05         & \underline{0.84}          & 2.74             & \underline{0.89}             \\
                              \bottomrule
\end{tabular}%
}
\end{table}

The VECtor dataset records scenes with both normal and fast motions. Due to the lower map quality under fast motion, we compare our method against EVT~\cite{10401990} (geometric method) and EVPT~\cite{bryner2019event} (photometric method) using the normal motion sequence. The methods for building the semi-dense map (required by EVT) and the dense map (required by EPVT) on VECtor dataset are based on the approach described in~\cite{yuan2024eviteventbasedvisualinertialtracking}. As not every method can track the entire sequence, the sequences are divided into 30\%, 50\%, and 100\% segments for a fair comparison. Given that EVT relies on a preset rate, we furthermore test it at different rates to determine the optimal performance. We use \emph{evo}~\cite{grupp2017evo} toolbox to align the first pair of the synchronized camera and ground truth poses and calculate RMSE as the average trajectory error (ATE) metric.

From Tab.~\ref{tab:vector_dataset}, it is evident that, although both our method and EVPT are photometric methods, our method is 2 to 10 times more accurate than EVPT. We are furthermore on par with the geometric method EVT, especially on the  sequence \emph{sofa\_normal}, where the event camera motion is relatively aggressive. EVPT and low-rate EVT are unable to complete the full sequence, whereas our method successfully does.
Unlike the sequence \emph{sofa\_normal}, which features rich appearance attributes with stuffed toys, it can be easily noted that the sequence \emph{desk\_normal} possesses sharper geometric features (e.g. desks, computers, and books). These features are generally advantageous for geometric methods like EVT. Nevertheless, EVT's performance is not always superior in such scenes, particularly when encountering occlusions (we will further demonstrate such effects on our self-collected dataset sequence \emph{desk}). The sequence \emph{robot\_normal} showcases a robot placed on a desk and again represents a scenario with smooth textures and clear appearance boundaries, thus slightly favouring the geometric approach over ours. In the continuation, we will analyze further self-collected sequences exhibiting more complex textures (e.g. a keyboard).

\begin{figure}[t]
    \centering
    \includegraphics[width=1\linewidth]{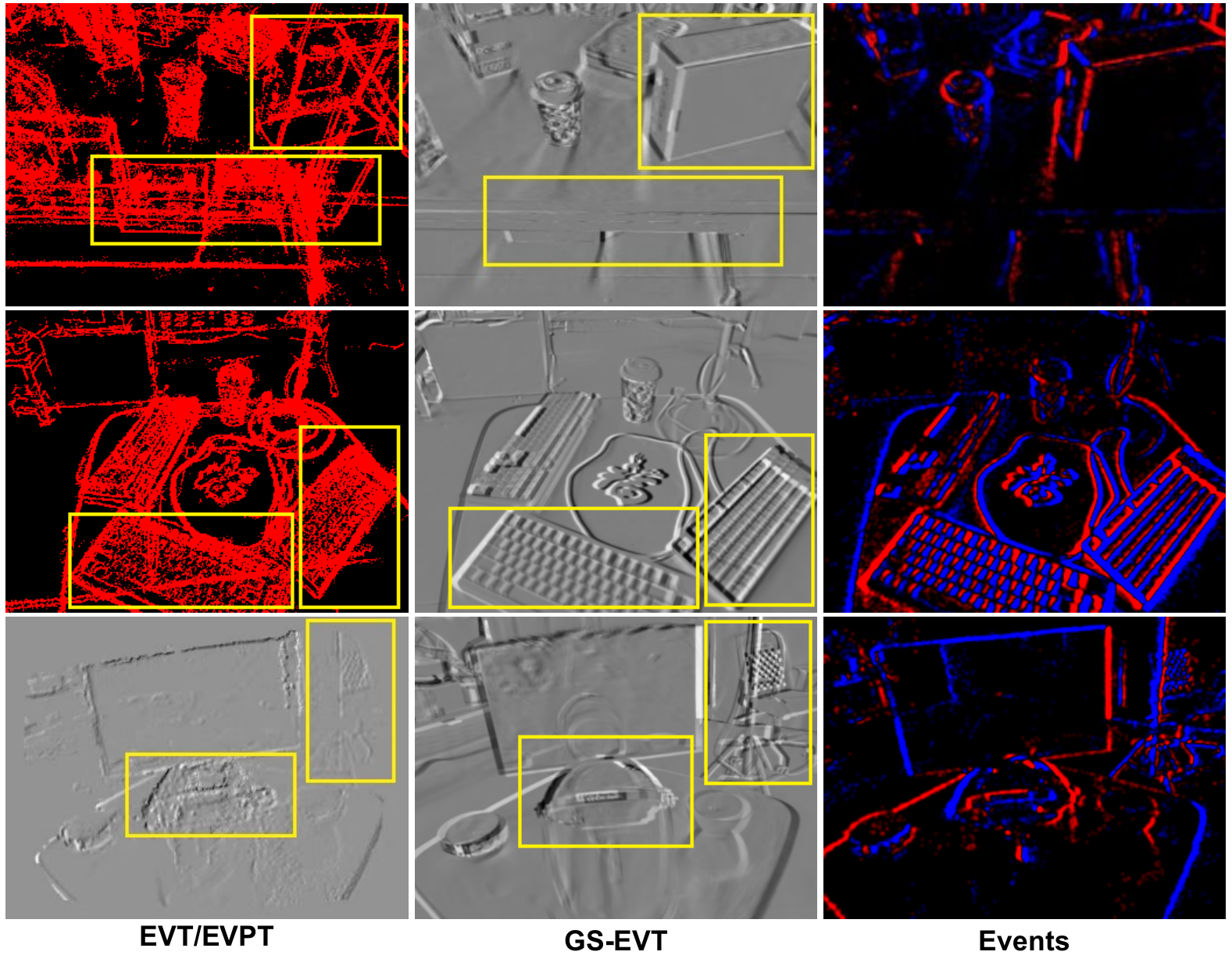}
    \caption{First row: Severely occluded scene type of sequence \emph{desk}. Reprojected semi-dense point cloud from EVT (left). Second row: Complexly textured scene type of sequence \emph{keyboard}. Reprojected semi-dense point cloud from EVT (left). Third row: Highly reflective scene type of  sequence \emph{helmet}. Rendered intensity change from EVPT (left).}
    \label{fig:comparsion}
\end{figure}

\subsection{Evaluation on self-collected dataset}

The effectiveness of a tracking algorithm significantly relies on the quality of the map it uses. To fully unlock the algorithm's potential, we propose our self-collected datasets in which map-building trajectories and tracking trajectories are separately recorded. By using separate trajectories, the quality of the map can be ensured without affecting the diversity of the evaluated tracking trajectories. We use a DAVIS346 to collect event streams for tracking and an iPad Pro to collect the color images for map-building. To ensure highly accurate ground truth trajectories, all sequences are captured under an OptiTrack motion capture system. To better demonstrate the algorithm's generalization ability, we record three scenes: \emph{desk} with boxes piled up and down leading to severe occlusions; \emph{keyboard} with tightly packed keys representing a noisy texture scenario; and \emph{helmet} exhibiting highly specular reflections as commonly encountered in man-made environments (cf. Fig.~\ref{fig:self-collected-dataset}).

\begin{figure}[h]
    \centering
    \includegraphics[width=0.9\linewidth]{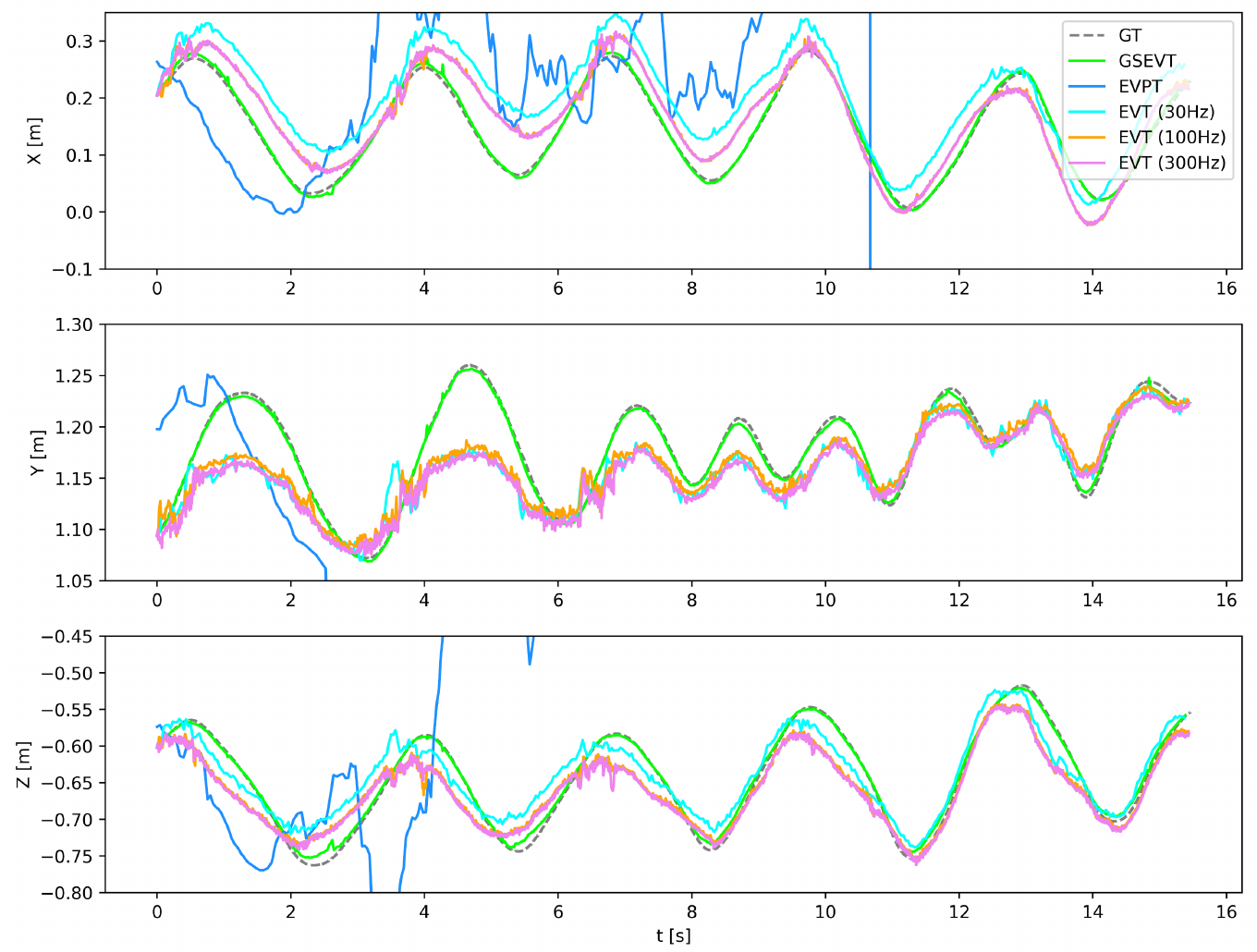}
    \caption{Comparison of translation error on \emph{helmet} sequence.}
    \label{fig:evo_traj_translation}
\end{figure}

Tab.~\ref{tab:self_collected_dataset} shows that our algorithm maintains high accuracy and outperforms EVT and EVPT methods under more challenging conditions. (To ensure that the superiority of our algorithm is not accidental, we collected four sequences for each scene, and the full results are provided in the supplementary materials.) The first row of Fig.~\ref{fig:comparsion} demonstrates that---although \emph{desk\_normal} from VECtor resembles our \emph{desk\_seq4} sequence---our self-collected dataset introduces additional challenges. For example, the occlusions created by stacked boxes easily challenges methods like EVT, which rely on semi-dense maps capturing only edge information. This potentially allows background edges to appear on foreground faces. What's more, semi-dense maps also struggle with complex textures. As seen in the sequence \emph{keyboard\_seq4} (cf. second row of Fig.~\ref{fig:comparsion}), EVT fails to capture the details of a keyboard, while our method continues to operate reliably. Although both EVPT and our method use dense maps, EVPT's discrete color point clouds can result in missing or blurry details. In contrast, our method employs 3DGS and thereby offers a more photo-realistic representation.
As further demonstrated on the sequence \emph{helmet\_seq4} (cf. third row of Fig.~\ref{fig:comparsion}), this also leads to better handling of complex lighting and reflections. In a nutshell, leveraging the powerful 3DGS representation and a coarse-to-fine pose and velocity optimization scheme, our method attains high ATE accuracy and smooth trajectory results across a large variety of scenarios (cf. Fig.~\ref{fig:evo_traj_translation}).

\begin{table}[t]
  \caption{Absolute trajectory error (ATE) on self-collected sequences. Position: {[}cm{]}, Orientation: {[}$^{\circ}${]} }
  \label{tab:self_collected_dataset}
  \resizebox{\columnwidth}{!}{%
  \begin{tabular}{@{}ccccccccc@{}}
  \toprule
  \multirow{2}{*}{Sequence} &
    \multirow{2}{*}{Method} &
    \multirow{2}{*}{Frequency} &
    \multicolumn{2}{c}{30\% sequence} &
    \multicolumn{2}{c}{50\% sequence} &
    \multicolumn{2}{c}{100\% sequence} \\ \cmidrule(l){4-9} 
                                &                      &       & Pos.          & Orient.       & Pos.          & Orient.       & Pos.          & Orient.       \\ \midrule
  \multirow{5}{*}{\makecell[c]{desk\_seq4 \\ (severe occlusion)}} & \multirow{3}{*}{EVT} & 300hz & -          & -         & -         & -          & -         & -          \\
                                  &  & 100hz & -          & -         & -         & -          & -         & -          \\
                                  &  & 30hz & -          & -         & -         & -          & -         & -          \\
                                & EVPT                 & (dyn)   & \underline{3.11}         & \underline{1.50}          & \underline{4.15}          & \underline{2.17}          & \underline{8.31}                 & \underline{7.17}             \\
                                & GSEVT                 & (dyn)   & \textbf{1.22}         & \textbf{0.34}         &\textbf{1.92}         & \textbf{0.68}          & \textbf{3.67}             & \textbf{2.05}             \\
                                \midrule
  \multirow{5}{*}{\makecell[c]{keyboard\_seq4 \\ (complex texture)}} & \multirow{3}{*}{EVT} & 300hz & \underline{2.73}          & \underline{1.47}          & -         & -          & -         & -          \\
                                  & & 100hz & 4.79          & \textbf{1.28}          & -         & -          & -         & -          \\
                                  &  & 30hz & -          & -          & -         & -          & -         & -          \\
                                  
                                & EVPT                 & (dyn)   & 5.05         & 4.62          & \underline{5.02}          & \underline{4.87}          & \underline{6.15}             & \underline{6.77}             \\
                                & GSEVT                 & (dyn)   & \textbf{2.22}         & 1.56         &\textbf{2.12}         & \textbf{1.66}          & \textbf{1.40}             & \textbf{1.08}             \\
                                \midrule
  \multirow{5}{*}{\makecell[c]{helmet\_seq1 \\ (high reflection)}} & \multirow{3}{*}{EVT} & 300hz & \underline{5.81}          & \underline{0.84}          & \underline{5.60}          & 0.87          & 5.94         & \underline{0.80}          \\
                                  &  & 100hz & 5.89          & 0.88          & 5.68          & 0.88          & 5.98         & \underline{0.80}          \\
                                  &  & 30hz & 6.08          & 0.87          & 5.82          & \underline{0.81}          & \underline{5.74}         & 0.82          \\
                                & EVPT                 & (dyn)   & 16.18         & 10.27          & -          & -          & -             & -             \\
                                & GSEVT                 & (dyn)   & \textbf{0.97}         & \textbf{0.37}          & \textbf{0.99}         & \textbf{0.38}          & \textbf{1.09}             & \textbf{0.43}             \\
                                \midrule
  \end{tabular}%
  }
  \end{table}

\section{CONCLUSION}
\label{sec:conclusion}

A novel framework for tracking the 6-DOF pose of an event camera with respect to a Gaussian splatting map is proposed in this work. By utilizing the 3DGS technique, our method uses a more photo-realistic map representation that could render view-dependent intensity changes, aligning better with the working principles of event cameras. Our method does not depend on depth information to simulate event generations~\cite{bryner2019event}, and directly leverages first-order dynamics for local differential image rendering, making the system more efficient. Unlike previous works~\cite{9386209, zuo2022devo} that focus solely on pose optimization, our method also introduces velocity optimization to further enhance the performance under highly dynamic conditions.

\newpage
\onecolumn
\section{SUPPLEMENTARY}
\label{sec:Supplementary}

\noindent\textbf{Camera Pose Jacobian:}
From Eq. (7), we see that $T_{cw}^\prime$ is composed of three transformations: $T_{cw}^\prime \mu = T_1 T_2 T_3\mu$, where $\mu$ represents the position of the 3D Gaussian ellipsoid in the world frame. As Eq. (8) shows, the gradient of $T_{cw}^\prime$ is related to $\mu^\prime$ and $\Sigma^\prime$. Applying the chain rule, the derivatives for the chain's front part are derived in the original Gaussian Splatting paper. We only need to focus on the derivatives: $\frac{\partial (T_{cw}^\prime \mu)}{\partial T_{cw}^\prime}$ and $\frac{\partial R_{cw}}{\partial T_{cw}^\prime}$. To optimize the pose, we should calculate the derivatives of $T_{cw}^\prime\mu$ and $R_{cw}$ w.r.t. $T_2$. Since $T_2$ is in $SE(3)$, instead of taking the derivative of $T_2$ directly, We optimize pose by optimizing the increments $(\Delta{t}, \Delta\theta)$ of $T_2$. Thus, we use the Lie algebra $\xi$ of $T_2$ and apply a perturbation $\Delta{T_2} = \text{exp}(\delta\xi^{\wedge})$ to $T_{cw}^\prime \mu$:

\begin{eqnarray}
    \begin{aligned}
        \frac{\partial(T_{cw}^\prime\mu)}{\partial(\Delta{t}, \Delta\theta)}
        &=\frac{\partial(T_{cw}^\prime\mu)}{\partial\delta\xi} \\
        &=\lim_{\delta\xi\to{0}} \frac{T_{1}\exp(\delta\xi^{\wedge})\exp(\xi^{\wedge})T_{3}\mu - T_{1}\exp(\xi^{\wedge})T_{3}\mu}{\delta\xi} \\
        &=\lim_{\delta\xi\to{0}} \frac{T_{1}(\mathbf{I}+\delta\xi^{\wedge})\exp(\xi^{\wedge})T_{3}\mu - T_{1}\exp(\xi^{\wedge})T_{3}\mu}{\delta\xi} \\
        &=\lim_{\delta\xi\to{0}} \frac{T_{1}\delta\xi^{\wedge}\exp(\xi^{\wedge})T_{3}\mu}{\delta\xi} \\
        &=\lim_{\delta\xi\to{0}} \frac{
        \begin{bmatrix}
            R_{1} & t_{1} \\
            \mathbf{0}^\mathrm{T} & 1
        \end{bmatrix}
        \begin{bmatrix}
            \delta\phi^{\wedge} & \delta\rho \\
            \mathbf{0}^\mathrm{T} & 0
        \end{bmatrix}
        \begin{bmatrix}
            R_{2} & t_{2} \\
            \mathbf{0}^\mathrm{T} & 1
        \end{bmatrix}
        \begin{bmatrix}
            R_{3} & t_{3} \\
            \mathbf{0}^\mathrm{T} & 1
        \end{bmatrix}
        \begin{bmatrix}
            \mu \\
            1
        \end{bmatrix}
        }{\delta\xi} \\
        &=\lim_{\delta\xi\to{0}} \frac{
        \begin{bmatrix}
            R_{1}\delta\phi^{\wedge}R_{2}R_{3}\mu + R_{1}\delta\phi^{\wedge}R_{2}t_{3} + R_{1}\delta\phi^{\wedge}t_{2} + R_{1}\delta\rho \\
            \mathbf{0}^\mathrm{T}
        \end{bmatrix}
        }{\delta\xi} \\
        &=\lim_{\delta\xi\to{0}} \frac{
        \begin{bmatrix}
            -R_{1}(R_{2}R_{3}\mu)^{\wedge}\delta\phi - R_{1}(R_{2}t_{3})^{\wedge}\delta\phi - R_{1}t_{2}^{\wedge}\delta\phi + R_{1}\delta\rho \\
            \mathbf{0}^\mathrm{T}
        \end{bmatrix}
        }{\begin{bmatrix}
            \delta\rho & \delta\phi
        \end{bmatrix}^\mathrm{T}} \\
        &= R_{1}
        \begin{bmatrix}
            \mathbf{I} & -[R_{2}(R_{3}\mu + t_{3}) + t_{2}]^{\wedge} \\
            \mathbf{0}^\mathrm{T} & \mathbf{0}^\mathrm{T}
        \end{bmatrix}
    \end{aligned}
\end{eqnarray}

Due to $R_{cw}$ only contains a rotation part, the derivative of $R_{cw}$ w.r.t. $\Delta t$ should be:

\begin{equation}
    \frac{\partial{R}_{cw}}{\partial\Delta{t}} =\begin{bmatrix}
            \mathbf{0} & \mathbf{0} & \mathbf{0}
        \end{bmatrix}^\mathrm{T}
\end{equation}

So we only need to consider the derivative of $R_{cw}$ w.r.t. $\Delta \theta$:

\begin{eqnarray}
    \begin{aligned}
        \frac{\partial R_{cw}}{\partial\Delta\theta}
        &=\displaystyle\lim_{\Delta\theta\to 0}
        \frac{R_{1}exp(\Delta\theta^{\wedge})R_{2}R_{3}-R{1}R_{2}R_{3}}{\Delta\theta}\\
        &=\displaystyle\lim_{\Delta\theta\to 0}
        \frac{R_{1}(\mathbf{I}+\Delta\theta^{\wedge})R_{2}R_{3}-R_{1}R_{2}R_{3}}{\Delta\theta}\\
        &=\displaystyle\lim_{\Delta\theta\to 0}
        R_{1}\frac{\Delta\theta^{\wedge}}{\Delta\theta}R_{2}R_{3}
    \end{aligned}
\end{eqnarray}

By differentiating one component (e.g., $\Delta\theta_x$) of $\Delta \theta^\wedge$, we can get:
\begin{equation}
\Delta\theta^\wedge = \begin{bmatrix}
  0&  -\Delta\theta_z& \Delta\theta_y\\
  \Delta\theta_z&  0& -\Delta\theta_x\\
  -\Delta\theta_y&  \Delta\theta_x& 0
\end{bmatrix} \to
\frac{\partial \Delta\theta^{\wedge}}{\Delta\theta_x} 
= \begin{bmatrix}
  0&  0& 0\\
  0&  0& -1\\
  0&  1& 0
\end{bmatrix}
\end{equation}

For each component of the partial derivative $\frac{\partial R_{cw}}{\partial\Delta\theta}$, we can get:
\begin{equation}
    \frac{\partial R_{cw}}{\partial\Delta\theta_x}
    = R_1 \begin{bmatrix}
  0&  0& 0\\
  0&  0& -1\\
  0&  1& 0
\end{bmatrix} R_2 R_3, \quad
    \frac{\partial R_{cw}}{\partial\Delta\theta_y}
    = R_1 \begin{bmatrix}
  0&  0& 1\\
  0&  0& 0\\
  -1&  0& 0
\end{bmatrix} R_2 R_3, \quad
    \frac{\partial R_{cw}}{\partial\Delta\theta_z}
    = R_1 \begin{bmatrix}
  0&  -1& 0\\
  1&  0& 0\\
  0&  0& 0
\end{bmatrix} R_2 R_3
\end{equation}

By denoting $(A)_{:, i}$ as the ith column of the matrix $A$, it can be simplified as follow:

\begin{equation}
    \frac{\partial{R}_{cw}}{\partial(\Delta{t}, \Delta\theta)} =R_{1}\begin{bmatrix}
            \mathbf{0} & -[(R_2R_3)_{:, 1}]^{\wedge}&   \\
             \mathbf{0} & -[(R_2R_3)_{:, 2}]^{\wedge}&   \\
             \mathbf{0} & -[(R_2R_3)_{:, 3}]^{\wedge}&   
        \end{bmatrix}
\end{equation}

\noindent\textbf{Camera Velocity Jacobian:} To optimize the velocity, we should calculate the derivatives of $T_{cw}^\prime\mu$ and $R_{cw}$ w.r.t. $T_1$. Since $[v, \omega]$ associates with time $\frac{\Delta\tau}{2}$ to represent transformation $T_1$, we need to further add the derivative $\frac{\partial (v \frac{\Delta\tau}{2}, \omega \frac{\Delta\tau}{2})}{\partial (v, \omega)} = [\frac{\Delta\tau}{2} \cdot \mathbf{I}, \frac{\Delta\tau}{2} \cdot \mathbf{I}]^\mathrm{T}$ following the chain rule. The subsequent steps follow a process similar to the differentiation of the camera pose Jacobian, so we present the results directly:

\begin{eqnarray}
\frac{\partial T_{cw}^\prime\mu}{\partial(v, \omega)} =
\frac{\Delta\tau}{2}
\begin{bmatrix}
    \mathbf{I}  &  -[R_{1}(R_{2}(R_{3}\mu+t_{3})+t_{2})+t_{1}]^{\wedge} \\
    \mathbf{0}^\mathrm{T}     &  \mathbf{0}^\mathrm{T}
\end{bmatrix}
\end{eqnarray}

\begin{eqnarray}
\frac{\partial R_{cw}}{\partial (v, \omega)} = 
\frac{\Delta\tau}{2}\begin{bmatrix}
     \mathbf{0} & -[(R_1R_2R_3)_{:, 1}]^{\wedge}   \\
     \mathbf{0} & -[(R_1R_2R_3)_{:, 2}]^{\wedge}   \\
     \mathbf{0} & -[(R_1R_2R_3)_{:, 3}]^{\wedge}
\end{bmatrix}
\end{eqnarray}

\noindent\textbf{Self-collected Dataset:} We list comparison results of all sequences in our self-collected dataset here (to make the table concise, we only show the general best frequency (300hz) of EVT): 

\begin{table}[htbp]
\centering
\caption{Absolute trajectory error (ATE) on self-collected sequences. Position: {[}cm{]}, Orientation: {[}$^{\circ}${]} }
\label{tab:normal}
\resizebox{0.5\columnwidth}{!}{%
\begin{tabular}{@{}ccccccccc@{}}
\toprule
\multirow{2}{*}{Sequence} &
  \multirow{2}{*}{Method} &
  \multirow{2}{*}{Frequency} &
  \multicolumn{2}{c}{30\% sequence} &
  \multicolumn{2}{c}{50\% sequence} &
  \multicolumn{2}{c}{100\% sequence} \\ \cmidrule(l){4-9} 
                              &                      &       & Pos.          & Orient.       & Pos.          & Orient.       & Pos.          & Orient.       \\ \midrule
\multirow{5}{*}{desk\_seq1} & EVT & 300hz & -          & -         & -         & -          & -         & -          \\
                              & EVPT                 & (dyn)   & -         & -          & -          & -          & -                 & -             \\
                              & GSEVT                 & (dyn)   & \textbf{0.70}         & \textbf{0.64}         &\textbf{0.80}         & \textbf{0.68}          & \textbf{0.98}             & \textbf{0.72}             \\
                              \midrule
\multirow{5}{*}{desk\_seq2} & EVT & 300hz & -          & -         & -         & -          & -         & -          \\
                              & EVPT                 & (dyn)   & -         & -          & -          & -          & -                 & -             \\
                              & GSEVT                 & (dyn)   & \textbf{1.03}         & \textbf{0.92}         &\textbf{1.11}         & \textbf{0.88}          & \textbf{1.30}             & \textbf{0.91}             \\
                              \midrule
\multirow{5}{*}{desk\_seq3} & EVT & 300hz & -          & -         & -         & -          & -         & -          \\
                              & EVPT                 & (dyn)   & -         & -          & -          & -          & -                 & -             \\
                              & GSEVT                 & (dyn)   & \textbf{1.28}         & \textbf{0.65}         &\textbf{2.60}         & \textbf{1.43}          & \textbf{4.47}             & \textbf{2.58}             \\
                              \midrule
\multirow{5}{*}{desk\_seq4} & EVT & 300hz & -          & -         & -         & -          & -         & -          \\
                              & EVPT                 & (dyn)   & 3.11         & 1.50          & 4.15          & 2.17          & 8.31                 & 7.17             \\
                              & GSEVT                 & (dyn)   & \textbf{1.22}         & \textbf{0.34}         &\textbf{1.92}         & \textbf{0.68}          & \textbf{3.67}             & \textbf{2.05}             \\
                              \midrule
\multirow{5}{*}{keyboard\_seq1} & EVT & 300hz & 1.84          & 0.75          & 1.66          & 0.79          & -         & -          \\
                              & EVPT                 & (dyn)   & 1.28        & 1.07         & 1.32          & 1.11          & 1.50             & 1.34             \\
                              & GSEVT                 & (dyn)   & \textbf{0.66}         & \textbf{0.50}         &\textbf{0.67}         & \textbf{0.47}          & \textbf{0.79}             & \textbf{0.59}             \\
                              \midrule
\multirow{5}{*}{keyboard\_seq2} & EVT & 300hz & -         & -          & -         & -          & -         & -          \\
                              & EVPT                 & (dyn)   & 6.29         & 8.96          & 5.81          & 8.48          & 5.78             & 8.38             \\
                              & GSEVT                 & (dyn)   & \textbf{0.88}         & \textbf{0.64}         &\textbf{0.95}         & \textbf{0.68}          & \textbf{1.03}             & \textbf{0.70}             \\
                              \midrule
\multirow{5}{*}{keyboard\_seq3} & EVT & 300hz & 1.17          & \textbf{0.80}          &-          & -       & -       & -          \\
                              & EVPT                 & (dyn)   & 2.14        & 2.15          & 4.03          & 2.30          & 4.34             & 2.22             \\
                              & GSEVT                 & (dyn)   & \textbf{1.08}         & 1.00         &\textbf{2.21}         & \textbf{1.8}          & \textbf{2.68}             & \textbf{1.92}             \\
                              \midrule
\multirow{5}{*}{keyboard\_seq4} & EVT & 300hz & 2.73          & 1.47          & -         & -          & -         & -          \\
                              & EVPT                 & (dyn)   & 5.05         & 4.62          & 5.02          & 4.87          & 6.15             & 6.77             \\
                              & GSEVT                 & (dyn)   & \textbf{2.22}         & 1.56         &\textbf{2.12}         & \textbf{1.66}          & \textbf{1.40}             & \textbf{1.08}             \\
                              \midrule
\multirow{5}{*}{helmet\_seq1} & EVT & 300hz & 5.81          & 0.84          & 5.60          & 0.87          & 5.94         & 0.80          \\
                              & EVPT                 & (dyn)   & 16.18         & 10.27          & -          & -          & -             & -             \\
                              & GSEVT                 & (dyn)   & \textbf{0.97}         & \textbf{0.37}          & \textbf{0.99}         & \textbf{0.38}          & \textbf{1.09}             & \textbf{0.43}             \\
                              \midrule
\multirow{5}{*}{helmet\_seq2} & EVT & 300hz & 7.11          & 1.08          & 7.76          & 0.93          & 6.16         & 0.67          \\
                              & EVPT                 & (dyn)   & -         & -          & -          & -          & -             & -             \\
                              & GSEVT                 & (dyn)   & \textbf{0.84}         & \textbf{0.42}          & \textbf{0.96}         & \textbf{0.47}          & \textbf{0.91}             & \textbf{0.41}             \\
                              \midrule
\multirow{5}{*}{helmet\_seq3} & EVT & 300hz & 3.61          & \textbf{0.43}          & 4.66          & \textbf{0.60}          & 5.99         & 0.66          \\
                              & EVPT                 & (dyn)   & 7.50         & 7.82          & 5.99          & 5.26          & -             & -             \\
                              & GSEVT                 & (dyn)   & \textbf{1.08}         & 0.46          & \textbf{1.51}         & \textbf{0.60 }         & \textbf{1.52}             & \textbf{0.65}             \\
                              \midrule
\multirow{5}{*}{helmet\_seq4} & EVT & 300hz & 4.69          & 0.65          & 4.45          & 0.66          & 4.99         & 0.68          \\
                              & EVPT                 & (dyn)   & -         & -          & -          & -          & -             & -             \\
                              & GSEVT                 & (dyn)   & \textbf{0.82}         & \textbf{0.48}         &\textbf{0.83}         & \textbf{0.52}          & \textbf{0.95}             & \textbf{0.58}             \\
                              \midrule
\end{tabular}%
}
\end{table}

\addtolength{\textheight}{0cm}   

\twocolumn
{
\small
\bibliographystyle{Bib/IEEEtran}
\bibliography{Bib/my}
}

\end{document}